\documentclass[conference]{./support/ieeeconf}
\usepackage{times}
\usepackage{amsmath}
\usepackage{balance}
\usepackage[pdftex]{graphicx}
\usepackage{subfigure}
\usepackage{amsmath,amssymb,amsopn,amstext,amsfonts}
\usepackage{cancel}
\usepackage[space]{cite}
\usepackage{pdfsync}
\usepackage{balance}
\usepackage{color}
\usepackage{algorithm}
\usepackage{algorithmic}
\usepackage{url}
\usepackage{booktabs}
\usepackage{threeparttable}
\usepackage[linkcolor=black,citecolor=black,urlcolor=black,colorlinks=true]{hyperref}
\usepackage{multirow}
\usepackage{graphicx}
\IEEEoverridecommandlockouts

\DeclareMathOperator*{\argminA}{arg\,min}

\graphicspath{{./figures/}}
\DeclareGraphicsExtensions{.pdf,.png,.jpg,.eps,.PNG}
\title{\LARGE \bf RoadMap: A Light-Weight Semantic Map for Visual Localization \\ towards Autonomous Driving}
\author{Tong Qin, Yuxin Zheng, Tongqing Chen, Yilun Chen, and Qing Su
\thanks{All authors are with  IAS BU Smart Driving Product Dept, Huawei Technologies, Shanghai, China.
        {\tt\small \{qintong, zhengyuxin, chentongqing, chengyilun, suqing\}@huawei.com}.}
}

\begin{document}

\maketitle
\thispagestyle{empty}
\pagestyle{empty}

\begin{abstract}
Accurate localization is of crucial importance for autonomous driving tasks. 
Nowadays, we have seen a lot of sensor-rich vehicles (e.g. Robo-taxi) driving on the street autonomously, which rely on high-accurate sensors (e.g. Lidar and RTK GPS) and high-resolution map.
However, low-cost production cars cannot afford such high expenses on sensors and maps.
How to reduce costs?
How do sensor-rich vehicles benefit low-cost cars?
In this paper, we proposed a light-weight localization solution, which relies on low-cost cameras and compact visual semantic maps.
The map is easily produced and updated by sensor-rich vehicles in a crowd-sourced way.
Specifically, the map consists of several semantic elements, such as lane line, crosswalk, ground sign, and stop line on the road surface.
We introduce the whole framework of on-vehicle mapping, on-cloud maintenance, and user-end localization.
The map data is collected and preprocessed on vehicles.
Then, the crowd-sourced data is uploaded to a cloud server.
The mass data from multiple vehicles are merged on the cloud so that the semantic map is updated in time.
Finally, the semantic map is compressed and distributed to production cars, which use this map for localization.
We validate the performance of the proposed map in real-world experiments and compare it against other algorithms.
The average size of the semantic map is $36$ kb/km.
We highlight that this framework is a reliable and practical localization solution for autonomous driving.
\end{abstract}

\section{Introduction}
There is an increasing demand for autonomous driving in recent years. 
To achieve autonomous capability, various sensors are equipped on the vehicle, such as GPS, IMU, camera, Lidar, radar, wheel odometry, etc. 
Localization is a fundamental function of an autonomous driving system. 
High precision localization relies on high precision sensors and High-Definition maps (HD Map).
Nowadays, RTK-GPS and Lidar are two common sensors, which are widely used for centimeter-level localization.
RTK-GPS provides accurate global pose in the open area by receiving signals from satellites and ground stations.
Lidar captures point cloud of surrounding environments.
Through point cloud matching, the vehicle can be localized in HD-Map in the GPS-denied environment.
These methods are already used in robot-taxi applications in a lot of cities.

Lidar and HD Map based solutions are ideal for robot-taxi applications.
However, there are several drawbacks which limit its usage on general production cars.
First of all, the production car cannot burden the high cost of Lidar and HD maps.
Furthermore, the point cloud map consumes a lot of memory, which is also unaffordable for mass production.
HD map production consumes huge amounts of manpower.
It is hard to guarantee timely updating.
To overcome these challenges, methods that rely on low-cost sensors and compact maps should be exploited.

\begin{figure}
	\centering
	\includegraphics[width=0.45\textwidth]{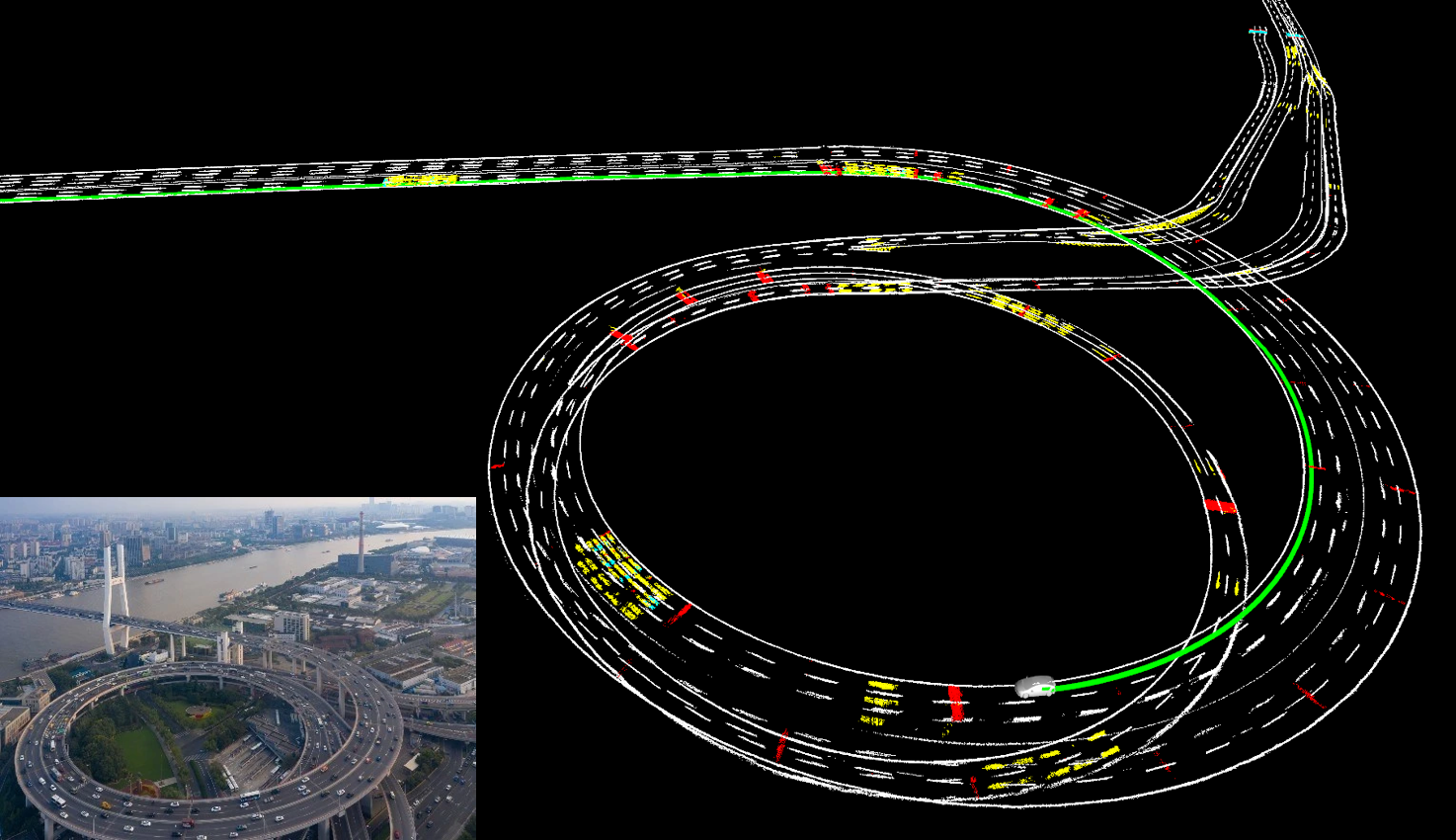}
	\caption{A sample picture at Nanpu Bridge, Shanghai.
		The semantic map contains lane lines (drawn in white) and other road markings (drawn in yellow and red).
		The green line is the vehicle's trajectory, which is localized based on this semantic map.
		The left-down picture is the real scene of the Nanpu Bridge captured in the birds-eye view.	  
	}
	\label{fig:abstract}
	\vspace{-1.0cm}
\end{figure}

\begin{figure*}
	\centering
	\includegraphics[width=0.9\textwidth]{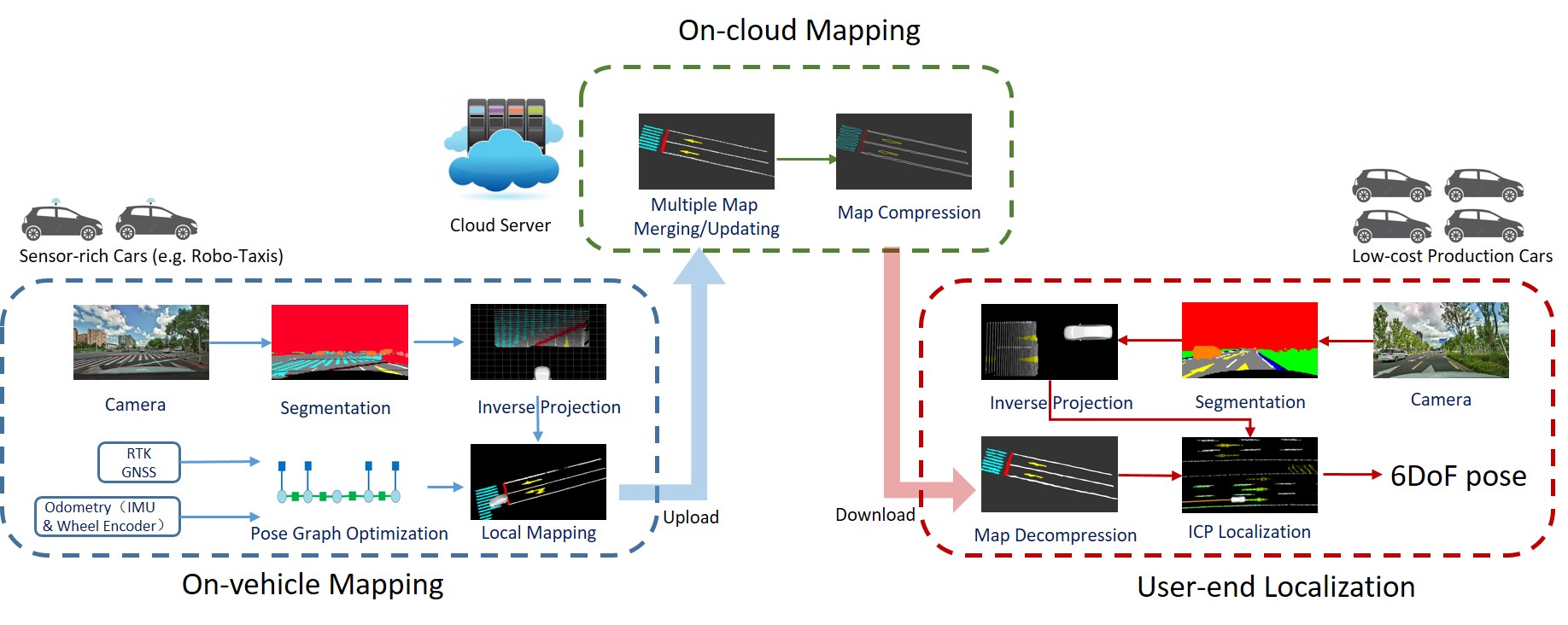}
	\caption{Illustration of the system structure. The system consists of three parts. The first part is the on-vehicle mapping. The second part is the on-cloud mapping. The last part is the end-user localization.}
	\label{fig:structure}
\end{figure*}

In this work, we present a light-weight localization solution, which relies on cameras and compact visual semantic maps. 
The map contains several semantic elements on the road, such as lane line, crosswalk, ground sign, and stop line.
This map is compact and easily produced by sensor-rich vehicles (e.g. Robo-taxis) in a crowd-sourced way.
The camera-Equipped low-cost vehicles can use this semantic map for localization.
To be specific, a learning-based semantic segmentation is used to extract useful landmarks. 
In the next, the semantic landmarks are recovered from 2D to 3D and registered into a local map.
The local map is uploaded to a cloud server.
The cloud server merges the data captured by different vehicles and compresses the global semantic map. 
Finally, the compact map is distributed to the production car for localization purposes. 
The proposed method of semantic mapping and localization is suitable for large-scale autonomous driving applications.
The contribution of this paper is summarized as follows:
\begin{itemize}
	\item We propose a novel framework for light-weight localization in autonomous driving tasks, which contains on-vehicle mapping, on-cloud map maintenance, and user-end localization.
	\item We propose a novel idea that uses sensor-rich vehicles (e.g. Robo-taxi) to benefit low-cost production cars, where sensor-rich vehicles collect data and update maps automatically every day.
	\item We conduct real-world experiments to validate the practicability of the proposed system.
\end{itemize}

\section{Literature Review}

Studies of visual mapping and localization became more and more popular over the last decades.
Traditional visual-based methods focus on Simultaneously Localization and Mapping (SLAM) in small-scale indoor environments.
In autonomous driving tasks, methods pay more attention to the large-scale outdoor environment. 

\subsection{Traditional Visual SLAM}
Visual Odometry (VO) is a typical topic in the visual SLAM area, which is widely applied in robotic applications.
Popular approaches include camera-only methods \cite{klein2007parallel, ForPizSca1405, engel2014lsd, mur2017orb, kitt2010visual} and visual-inertial methods \cite{MouRou0704, qin2018vins, LeuFurRab1306, forster2017manifold}.
Geometrical features such as sparse points, lines, and dense planes in the natural environment are extracted.
Among these methods, corner feature points are used in \cite{klein2007parallel, mur2017orb, qin2018vins, LeuFurRab1306, LiMou1305}.
The camera pose, as long as feature positions, are estimated at the same time.
Features can be further described by descriptors to make them distinguishable.
Engel \textit{et al.}\cite{engel2014lsd} leveraged semi-dense edges to localize camera pose and build the environment's structure.

Expanding the visual odometry by adding a prior map, it becomes the localization problem with a fixed coordinate.
Methods, \cite{lynen2015get, mur2017orb, schneider2018maplab, burki2016appearance, QinShen18}, built a visual map in advance, then relocalized camera pose within this map.
To be specific, visual-based methods \cite{lynen2015get, burki2016appearance} localized camera pose against a visual feature map by descriptor matching. 
The map contains thousands of 3D visual features and their descriptors.   
Mur-Artal \textit{et al.} \cite{mur2017orb} leveraged ORB features \cite{rublee2011orb} to build a map of the environment.
Then the map can be used to relocalize the camera by ORB descriptor matching.
Moreover, methods\cite{schneider2018maplab, QinShen18}, can automatically merge multiple sequences into a global frame.
For autonomous driving tasks, Burki \textit{et al.} \cite{burki2016appearance} demonstrated the vehicle was localized by the sparse feature map on the road.
Inherently, traditional appearance-based methods were suffered from light, perspective, and time changes in the long term. 

\subsection{Road-based Localization}
Road-based localization methods take full advantage of road features in autonomous driving scenarios.
Road features contain various markers on the road surface, such as lane lines, curbs, and ground signs.
Road features also contain 3D elements, such as traffic lights and traffic signs. 
Compared with traditional features, these markings are abundant and stable on the road and robust against time and illumination changes.
In general, an accurate prior map (HD map) is necessary.
This prior map is usually built by high-accurate sensor setups (Lidar, RTK GNSS, etc.).
The vehicle is localized by matching visual detection with this map. 
To be specific, Schreiber \textit{et al.} \cite{schreiber2013laneloc} localized camera by detecting curbs and lanes, and matching the structure of these elements with a highly accurate map.
Further on, Ranganathan \textit{et al.} \cite{ranganathan2013light} utilized road markers and detected corner points on road markers.
These key points were used for localization.
Yan \textit{et al.} \cite{lu2017monocular} formulated a non-linear optimization to match the road markings with the map.
The vehicle odometry and epipolar geometry were also taken into consideration to estimate the 6-DoF camera pose.

Meanwhile, some researches \cite{rehder2015submap, jeong2017road,  qin2020avp, herb2019crowd} focused on building the road map.
Regder \textit{et al.} \cite{rehder2015submap} detected lanes on the image and used odometry to generate local grid maps.
The pose was further optimized by local map stitching.
Moreover, Jeong \textit{et al.} \cite{jeong2017road} classified road markings.
Informative classes were used to avoid ambiguity.
Loop closure and pose graph optimization were performed to eliminate drift and maintain global consistency.
Qin \textit{et al.} \cite{qin2020avp} built the semantic map of underground parking lots by road markers.
The autonomous parking application was performed based on this semantic map.
Herb \textit{et al.} \cite{herb2019crowd} proposed a crow-sourced way to generate semantic map.
However, it was difficult to be applied since the inter-session feature matching consumed great computation.

In this paper, we focus on a complete system that contains on-vehicle mapping, on-cloud map merging / updating, and end-user localization.
The proposed system is a reliable and practical localization solution for large-scale autonomous driving.

\begin{figure}
	\centering
	\subfigure[Raw image.]{
		\label{fig:raw_image}    
		\includegraphics[width=0.23\textwidth]{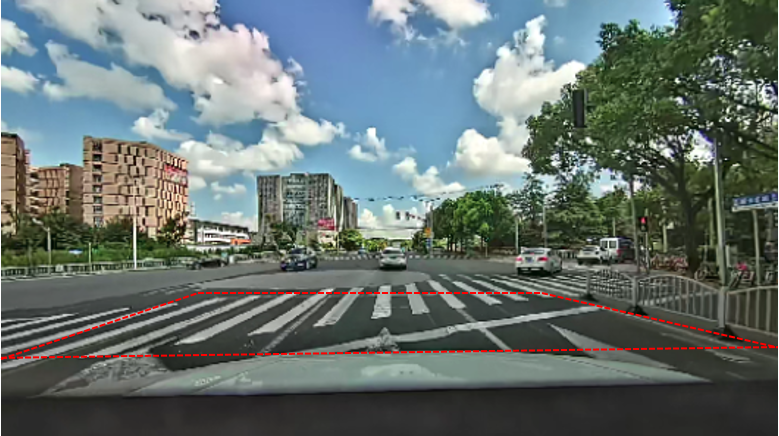}}   
	\subfigure[Segmentation image.]{
		\label{fig:semantic}
		\includegraphics[width=0.23\textwidth]{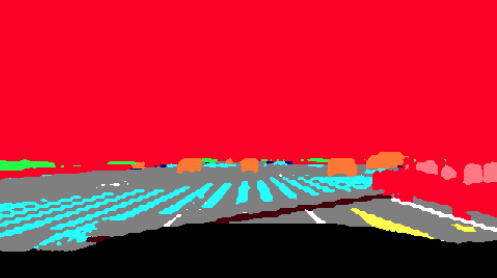}}
		\subfigure[Segmentic features under vehicle's coordinate.]{
		\label{fig:semantic_points}
		\includegraphics[width=0.23\textwidth]{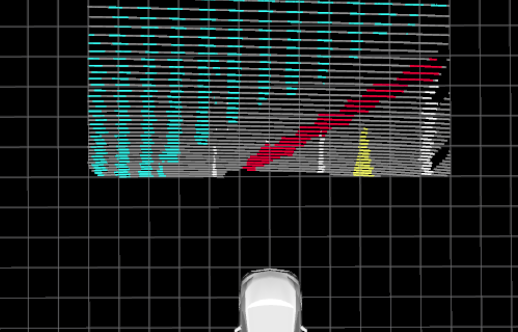}}
	\caption{(a) is the raw image captured by front-view camera. The red box is ROI used in Sec. \ref{sec:ipm}.  (b) is the segmentation result, which classifies the scene into multiple classes.
		The lane line is drawn in white. The crosswalk is drawn in blue. The road mark is drawn in yellow. The stop line is drawn in brown. The road surface is drawn in gray. (c) shows the semantic feature under the vehicle's coordinate.}
	\label{fig:segmentation}
\end{figure}

\section{System Overview}

The system consists of three parts, as shown in Fig. \ref{fig:structure}. 
The first part is the on-vehicle mapping.
Vehicles equipped with a front-view camera, RTK-GPS, and basic navigation sensors (IMU and wheel encoder) are used.
These vehicles are widely used for Robot-taxi applications, which collect a lot of real-time data every day.
Semantic features are extracted from the front-view image by a segmentation network. 
Then semantic features are projected to the world frame based on the optimized vehicle's pose.
A local semantic map is built on the vehicle.
This local map is uploaded to a cloud map server.

The second part is on-cloud mapping.
The cloud server collects local maps from multiple vehicles.
The local map is merged into a global map.
Then the global map is compressed by contour extraction.
Finally, the compressed semantic map is released to end-users.

The last part is the end-user localization.
The end users are production cars, which equip low-cost sensors, such as cameras, low-accurate GPS, IMU, and wheel encoders.
The end-user decodes the semantic map after downloading it from the cloud server.
Same as the on-vehicle mapping part, semantic features are extracted from the front-view image by segmentation.
The vehicle is localized against the map by semantic feature matching.

\section{On-Vehicle Mapping}

\subsection{Image Segmentation}
We utilize a CNN-based method, such as \cite{long2015fully, ronneberger2015u, badrinarayanan2015segnet}, for scene segmentation.
In this paper, the front-view image is segmented into multiple classes, such as ground, lane line, stop line, road marker, curb, vehicle, bike, and human.
Among these classes, ground, lane line, stop line, and road marker are used for semantic mapping.
Other classes can be used for other autonomous driving tasks.
An example of image segmentation is shown in Fig \ref{fig:segmentation}.
Fig \ref{fig:raw_image} shows the raw image captured by front-view camera.
Fig \ref{fig:semantic} shows the corresponding segmentation result.

\begin{figure}
	\centering
	\includegraphics[width=0.4\textwidth]{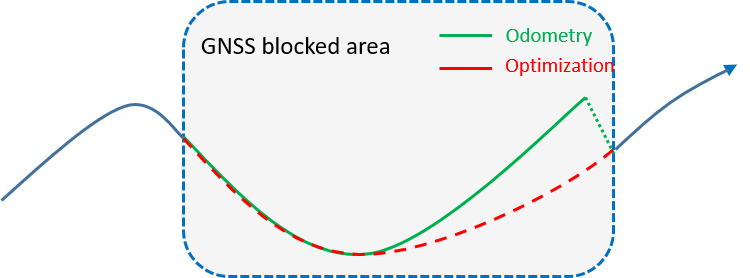}
	\caption{The blue line is the trajectory in GNSS-good area, which is accurate thanks to high-accurate RTK GNSS.
		In the GNSS-blocked area, the odometry trajectory is drawn in green, which drifts a lot. 
		To relieve drift, pose graph optimization is performed.
		The optimized trajectory is drawn in red, which is smooth and drift-free.}
	\label{fig:gnss_block}
	\vspace{-0.5cm}
\end{figure}

\subsection{Inverse Perspective Transformation}
\label{sec:ipm}

After segmentation, the semantic pixels are inversely projected from the image plane to the ground plane under the vehicle's coordinate.
This procedure is also called Inverse Perspective Mapping (IPM).
The intrinsic parameter of the camera and the extrinsic transformation from the camera to the vehicle's center are offline calibrated. 
Due to the perspective noise, the farther the scene, the greater the error is.
We only select pixels in a Region Of Interest (ROI), where is close to the camera center, as shown in Fig. \ref{fig:raw_image}.
This ROI denotes a $12m \times 8m$ rectangle area in front of the vehicle.
Assuming the ground surface is a flat plane, each pixel $[u, v]$ is projected into the ground plane (z equals 0) under the vehicle's coordinate as follows,

\begin{equation}
\label{eq:ipm}
\frac{1}{\lambda}
\begin{bmatrix}
x^v \\ y^v \\ 1
\end{bmatrix}
= 
\begin{bmatrix}
\mathbf{R}_c \ \mathbf{t}_c
\end{bmatrix}
_{col:1,2,4} ^ {-1}
\pi_c ^ {-1}(
\begin{bmatrix}
u\\v\\1
\end{bmatrix}
),
\end{equation}
where $\pi_c(\cdot)$ is the distortion and projection model of the camera.
$\pi_c(\cdot)^{-1}$ is the inverse projection, which lifts pixel into space. 
$[\mathbf{R}_c \ \mathbf{t}_c]$ is the extrinsic matrix of each camera with respect to vehicle's center.
$[u \ v]$ is pixel location in image coordinate.
$[x^v \ y^v]$ is the position of the feature in the vehicle's center coordinate.  
$\lambda$ is a scalar.
$()_{col:i}$ means taking the $i_{th}$ column of this matrix.
An example result of the inverse perspective transformation is shown in Fig. \ref{fig:semantic_points}.
Each labeled pixel in ROI is projected on the ground in front of the vehicle.

\begin{figure}
	\centering
	\includegraphics[width=0.4\textwidth]{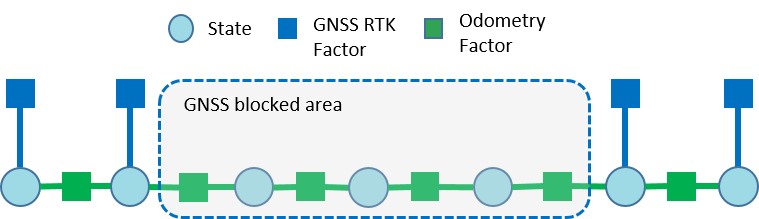}
	\caption{Illustration picture of the pose graph.
		The blue node is the state of the vehicle at a certain time, which contains position $\mathbf{p}_t$ and orientation $\mathbf{R}_t$.
		There are two kinds of edge.
		The blue edge denotes the GNSS constraint, which only exists in GNSS-good time.
		It affects only one node.
		The green edge denotes the odometry constraint, which exists at any time.
		It constrains two neighbor nodes.
	}
	\label{fig:pose_graph}
\end{figure}

\subsection{Pose Graph Optimization}

To build the map, the accurate pose of the vehicle is necessary.
Though RTK-GNSS is used, it can't guarantee a reliable pose all the time.
RTK-GNSS can provide centimeter-level position only in the open area.
Its signal is easily blocked by tall buildings in the urban scenario.
Navigation sensors (IMU and wheel) can provide odometry in the GNSS-blocked area.
However, odometry is suffered from accumulated drift for a long time.
An illustration picture of this problem is shown in Fig. \ref{fig:gnss_block}.
The blue line is the trajectory in the GNSS-good area, which is accurate thanks to high-accurate RTK GNSS.
In the GNSS-blocked area, the odometry trajectory is drawn in green, which drifts a lot. 
To relieve drift, pose graph optimization is performed.
The optimized trajectory is drawn in red, which is smooth and drift-free.

An illustration picture of the pose graph is shown in Fig. \ref{fig:pose_graph}.
The blue node is the state $\mathbf{s}$ of the vehicle at a certain time, which contains position $\mathbf{p}$ and orientation $\mathbf{q}$.
We use quaternion $\mathbf{q}$ to denote orientation.
The operation $\mathbf{R}(\mathbf{q})$ convert a quaternion to the rotation matrix.
There are two kinds of edge.
The blue edge denotes the GNSS constraint, which only exists in the GNSS-good moment.
It affects only one node.
The green edge is the odometry constraint, which exists at any time.
It constrains two neighbor nodes.
The pose graph optimization can be formulated as following equation:

\begin{equation}
\min_{\mathbf{s}_0...\mathbf{s}_n} \left\{ 
\sum_{i\in[1,n]}
\|
{\mathbf{r}_o(\mathbf{s}_{i-1}, \mathbf{s}_i, \hat{\mathbf{m}}^o_{i-1,i})}
 \|^2_{\boldsymbol{\sigma}}
 +
\sum_{i\in \mathcal{G}} 
\|
{\mathbf{r}_g(\mathbf{s_i}, \hat{\mathbf{m}}^g_i)}
\|^2_{\boldsymbol{\sigma}}
\right\},
\end{equation}
where $\mathbf{s}$ is the pose state (position and orientation).
$\mathbf{r}_o$ is the residual of odometry factor.
$\hat{\mathbf{m}}^o_{i-1, i}$ is the odometry measurement, which contains delta position $\delta\hat{\mathbf{p}}_{i-1, i}$ and orientation $\delta\hat{\mathbf{q}}_{i-1, i}$ between two neighbor states.
$\mathbf{r}_g$ is the residual of GNSS factor.
$\mathcal{G}$ is the set of states in GNSS-good aera.
$\hat{\mathbf{m}}^g_i$ is the GNSS measurement, which is the position $\hat{\mathbf{p}}_{i}$ in the global frame.
The residual factor $\mathbf{r}_o$ and $\mathbf{r}_g$ are defined as follows:
\begin{equation}
\begin{aligned}
\mathbf{r}_o(\mathbf{s}_{i-1}, \mathbf{s}_i, \hat{\mathbf{m}}^o_{i-1, i}) &= 
\begin{bmatrix}
{\mathbf{R}(\mathbf{q}_{i-1})}^{-1}(\mathbf{p}_i - \mathbf{p}_{i-1}) - \delta\hat{\mathbf{p}}^o_{i-1,i} \\
[{\mathbf{q}_{i}}^{-1} \cdot \mathbf{q}_{i-1} \cdot \delta\hat{\mathbf{q}}_{i-1, i}^o]_{xyz}
\end{bmatrix}
\\
\mathbf{r}_g(\mathbf{s_i}, \hat{\mathbf{m}}^g_i) &=
\mathbf{p}_i - \hat{\mathbf{m}}^g_i
\end{aligned},
\end{equation}
where, $[\cdot]_{xyz}$ takes the first three elements of quaternion, which approximately equals to an error perturbation on the manifold.

\begin{figure}
	\centering
		\subfigure[The semantic map.]{
		\label{fig:origin}    
		\includegraphics[width=0.38\textwidth]{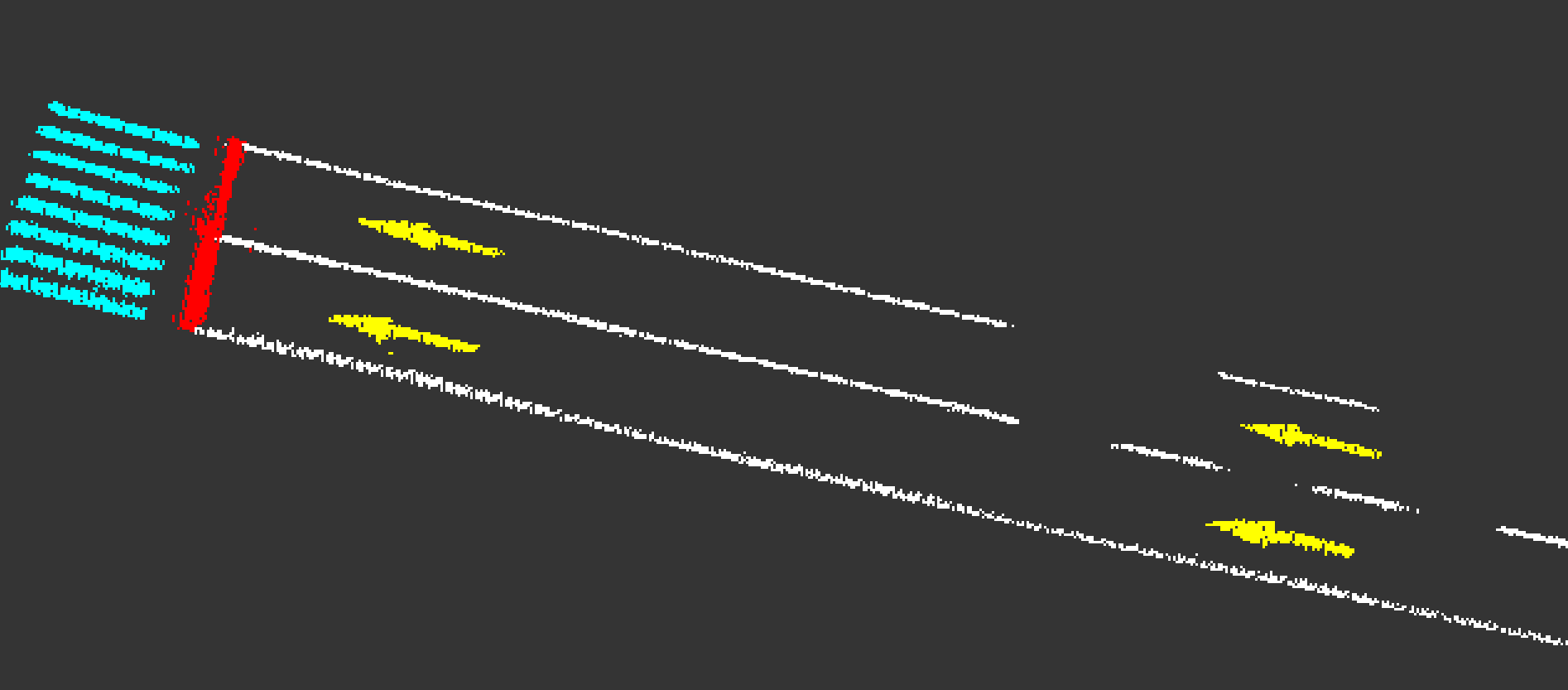}}  
	\subfigure[The contour of the semantic map.]{
		\label{fig:compact_2}    
		\includegraphics[width=0.38\textwidth]{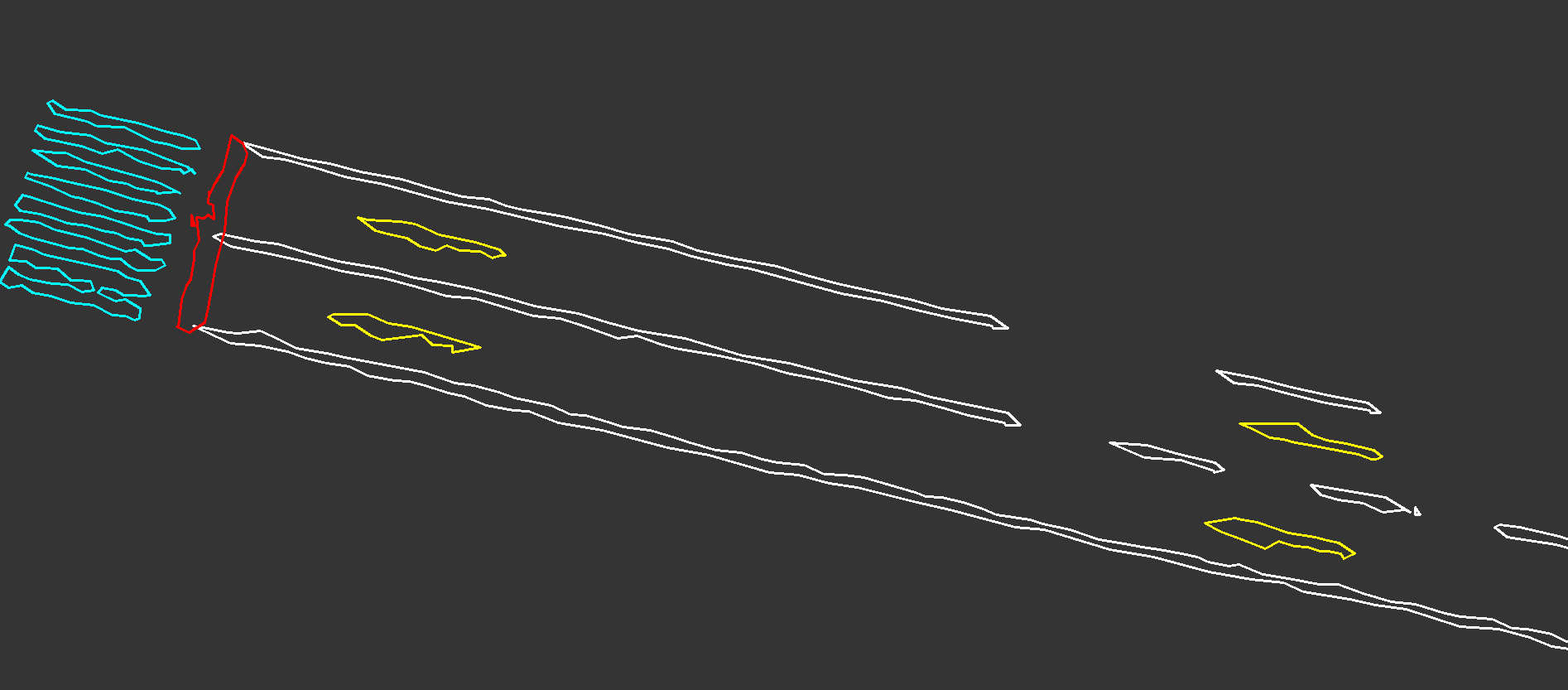}}   
	\subfigure[Recovered semantic map from the contour.]{
		\label{fig:recover}
		\includegraphics[width=0.38\textwidth]{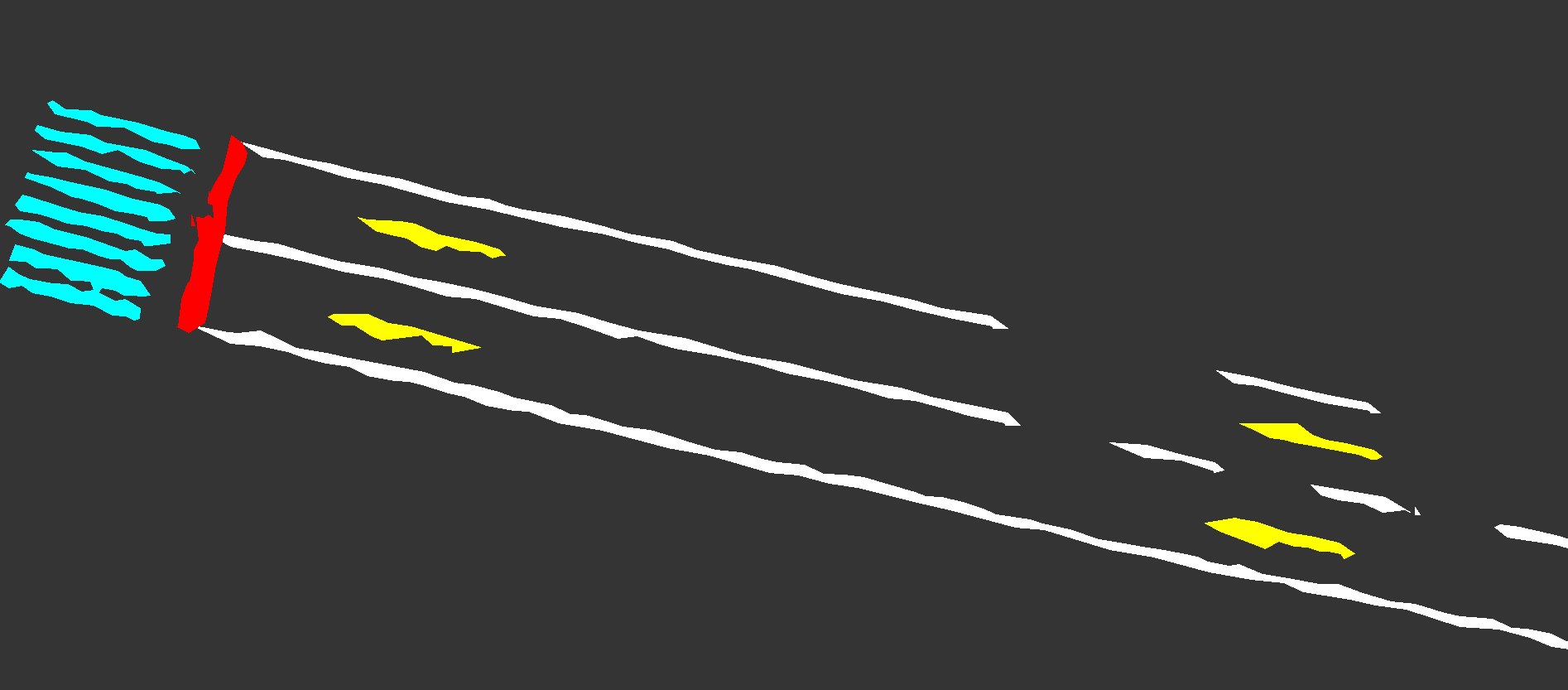}}
	\caption{A sample of semantic map compression and decompression. (a) shows the raw semantic map. (b) shows the contour of this semantic map. (c) shows the recovered semantic map form the contour.}
	\label{fig:compact_recover}
\end{figure}

\subsection{Local Mapping}
Pose graph optimization provides a reliable vehicle's pose at every moment.
The semantic features captured in frame $i$ are transformed from vehicle's coordinate into the global coordinate based on this optimized pose,

\begin{equation}
\label{eq:trans}
\begin{bmatrix}
x^w \\ y^w \\ z^w
\end{bmatrix}
= \mathbf{R}(\mathbf{q}_i) 
\begin{bmatrix}
x^v \\ y^v \\ 0
\end{bmatrix}
+ \mathbf{p}_i
.
\end{equation}
From image segmentation, each point contains a class label (ground, lane line, road sign, and crosswalk).
Each point presents a small area in the world frame.
When the vehicle is moving, one area can be observed multiple times.
However, due to the segmentation noise, this area may be classified into different classes.
To overcome this problem, we use statistics to filter noise. 
The map is divided into small grids, whose resolution is $0.1\times0.1\times0.1m$. 
Every grid's information contains position, semantic labels, and counts of each semantic label.
Semantic labels include ground, lane line, stop line, ground sign, and crosswalk. 
In the beginning, the score of each label is zero.
When a semantic point is inserted into a grid, the score of the corresponding label increases one. 
Thus, the semantic label with the highest score presents the grid's class. 
Through this method, the semantic map becomes accurate and robust to segmentation noise.
An example of global mapping results is shown in Fig. \ref{fig:origin}.

\section{On-Cloud Mapping}

\subsection{Map Merging / Updating}
A cloud mapping server is used to aggregate mass data captured by multiple vehicles.
It merges local maps timely so that the global semantic map is up-to-date.
To save the bandwidth, only the occupied grid of the local map is uploaded to the cloud.
Same as the on-vehicle mapping process, the semantic map on the cloud server is also divided into grids with a resolution of $0.1\times0.1\times0.1m$.
The grid of the local map is added to the global map, according to its position.
Specifically, the score in the local map's grid is added to the corresponding grid on the global map.
This process is operated in parallel. 
Finally, the label with the highest score is the grid's label.
A detailed example of map updating is shown in Fig. \ref{fig:map_update}.

\subsection{Map Compression}
Semantic maps generated in the cloud server will be used for localization for a large number of production cars. 
However, the transmission bandwidth and onboard storage are limited on the production car. 
To this end, the semantic map is further compressed on the cloud. 
Since the semantic map can be efficiently presented by the contour, we use contour extraction to compress the map. 
Firstly, we generate the top view image of the semantic map. 
Each pixel presents a grid. 
Secondly, the contour of every semantic group is extracted.
Finally, the contour points are saved and distributed to the production car.
As shown in Fig. \ref{fig:compact_recover}, (a) shows the raw semantic map. 
Fig. \ref{fig:compact_2} shows the contour of this semantic map.

\begin{figure}
	\centering
	\includegraphics[width=0.38\textwidth]{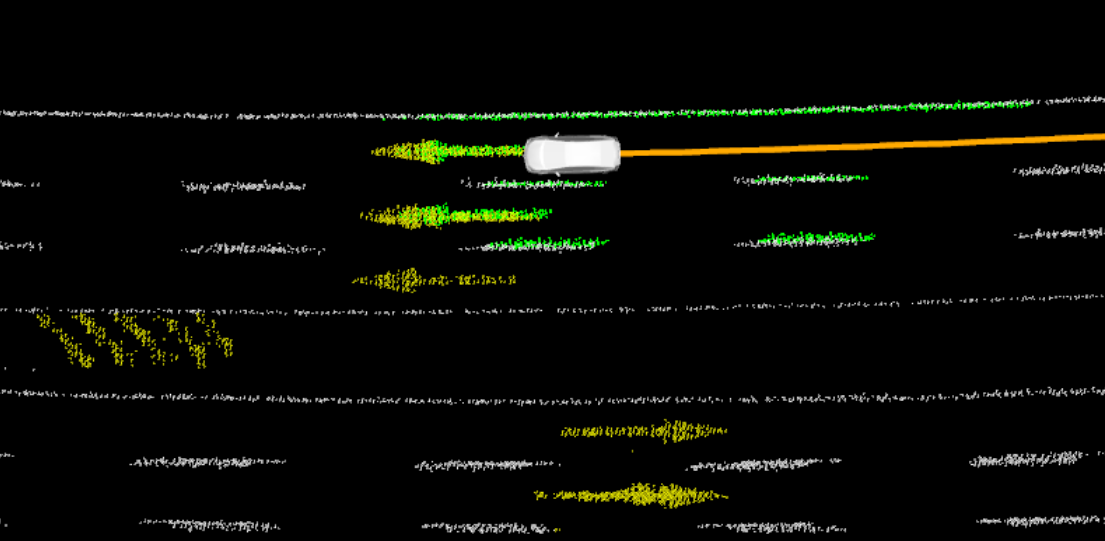}
	\caption{Illustration of localization in the semantic map. The white and yellow dots are lane lines and road markings on the map. The green dots are observed semantic features. The vehicle is localized by matching current features with the map. The orange line is the estimated trajectory.}
	\label{fig:loc}
\end{figure}

\section{User-end Localization}

The end user denotes production cars, which equip low-cost sensors, such as cameras, low-accurate GPS, IMU, and wheel encoders.

\subsection{Map Decompression}
When the end-user receives the compressed map, the semantic map is decompressed from contour points.  
In the top-view image plane, we fill points inside the contours with the same semantic labels.
Then every labeled pixel is recovered from the image plane into the world coordination.  
An example of decompressed semantic maps is shown in Fig. \ref{fig:recover}.
Comparing original map in Fig. \ref{fig:origin} with recovered map in Fig. \ref{fig:recover}, the decoder method efficiently recovers semantic information.

\subsection{ICP Localization}
This semantic map is further used for localization.
Similar to the mapping procedure, the semantic points are generated from the front-view image segmentation and projected into the vehicle frame.
Then the current pose of the vehicle is estimated by matching current feature points with the map, as shown in Fig. \ref{fig:loc}.
The estimation adopts the ICP method, which can be written as following equation,  
\begin{equation}
\mathbf{q}^*, \mathbf{p}^* = \argminA_{\mathbf{q},\mathbf{p}} \sum_{k\in \mathcal{S}} \| \mathbf{R}(\mathbf{q}) 
\begin{bmatrix}
x^v_k \\ y^v_k \\ 0
\end{bmatrix}
+ \mathbf{p}
-
\begin{bmatrix}
x^w_k \\ y^w_k \\ z^w_k
\end{bmatrix}
\| ^2
,
\end{equation}
where $\mathbf{q}$ and $\mathbf{p}$ are quaternion and position of the current frame.
$\mathcal{S}$ is the set of current feature points.
$[x^v_k \ y^v_k \ 0]$ is the current feature under the vehicle coordinate.
$[x^w_k \ y^w_k \ z^w_k]$ is the closest point of this feature in the map under global coordinate.

An EKF framework is adopted at the end, which fuses odometry with visual localization results.
The filter not only increases the robustness of the localization but also smooths the estimated trajectory.

\begin{figure}
	\centering
	\includegraphics[width=0.38\textwidth]{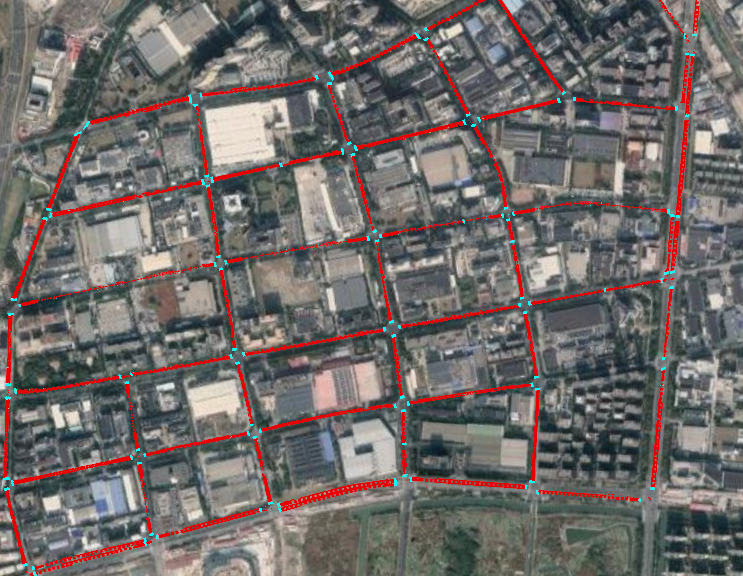}
	\caption{The semantic map of a urban block in Pudong New District, Shanghai. The map is aligned with Google Map.}
	\label{fig:exp}
\end{figure}

\section{Experimental Results}

We validated the proposed semantic map through real-world experiments.
The first experiment focused on mapping capability.
The second experiment focused on localization accuracy.

\subsection{Map Production}
Vehicles equipped with RTK-GPS, front-view camera, IMU, and wheel encoders were used for map production.
Multiple vehicles run in the urban area at the same time.
On-vehicle maps were uploaded onto the cloud server through the network.
The final semantic map was shown in Fig. \ref{fig:exp}.
The map covered an urban block in Pudong New District, Shanghai.
We aligned the semantic map with Google Map.
The overall length of the road network in this area was $22$ KM.
The whole size of the raw semantic map was $16.7$ MB.
The size of the compressed semantic map was $0.786$ MB.
The average size of the compressed semantic map was $36$ KB/KM.

\begin{figure}
	\centering
	\subfigure[The lane line before redrawn.]{
		\label{fig:raw0}    
		\includegraphics[width=0.23\textwidth]{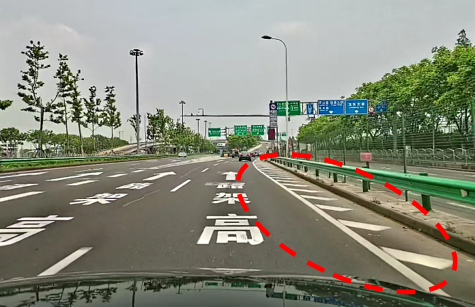}}   
	\subfigure[The lane line after redrawn.]{
		\label{fig:raw1}
		\includegraphics[width=0.23\textwidth]{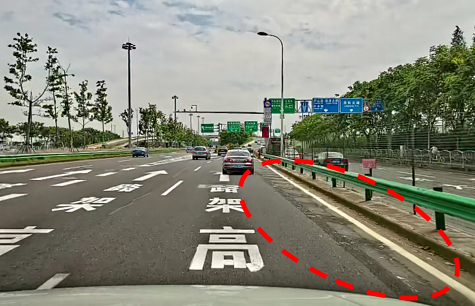}}
	\subfigure[Segmentation map before lane line was redrawn.]{
		\label{fig:map0}
		\includegraphics[width=0.4\textwidth]{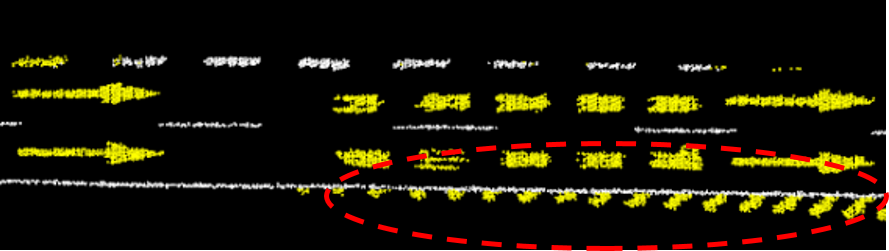}}
	\subfigure[Segmentation map were updating.]{
		\label{fig:map1}
		\includegraphics[width=0.4\textwidth]{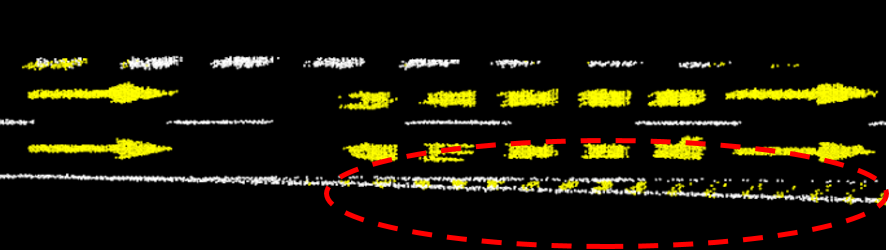}}
	\subfigure[Segmentation map after lane line was redrawn.]{
		\label{fig:map2}
		\includegraphics[width=0.4\textwidth]{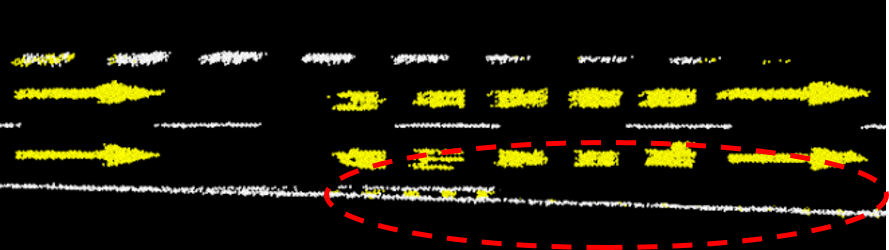}}
	\caption{An illustration of semantic map updating when environment changed. (a) shown the original environment. (b) shown the changed environment where lane line was redrawn. (c) shown the original semantic map. (d) shown the semantic map was updating, when the new lane lines were replacing the old. (e) shown the final semantic map. }
	\label{fig:map_update}
\end{figure}

A detailed example of map updating progress was shown in Fig. \ref{fig:map_update}.
Lane lines were redrawn in the area.
Fig. \ref{fig:raw0} shown the original lane line.
Fig. \ref{fig:raw1} shown the lane line after it was redrawn.
The redrawn area was highlighted in the red circle.
Fig. \ref{fig:map0} shown the original semantic map.
Fig. \ref{fig:map1} shown the semantic map was updating.
The new lane line was replacing the old one gradually.
With merging more and more up-to-date data, the semantic map was updated completely, as shown in Fig. \ref{fig:map2}.

\subsection{Localization Accuracy}

\begin{figure}
	\centering
	\includegraphics[width=0.48\textwidth]{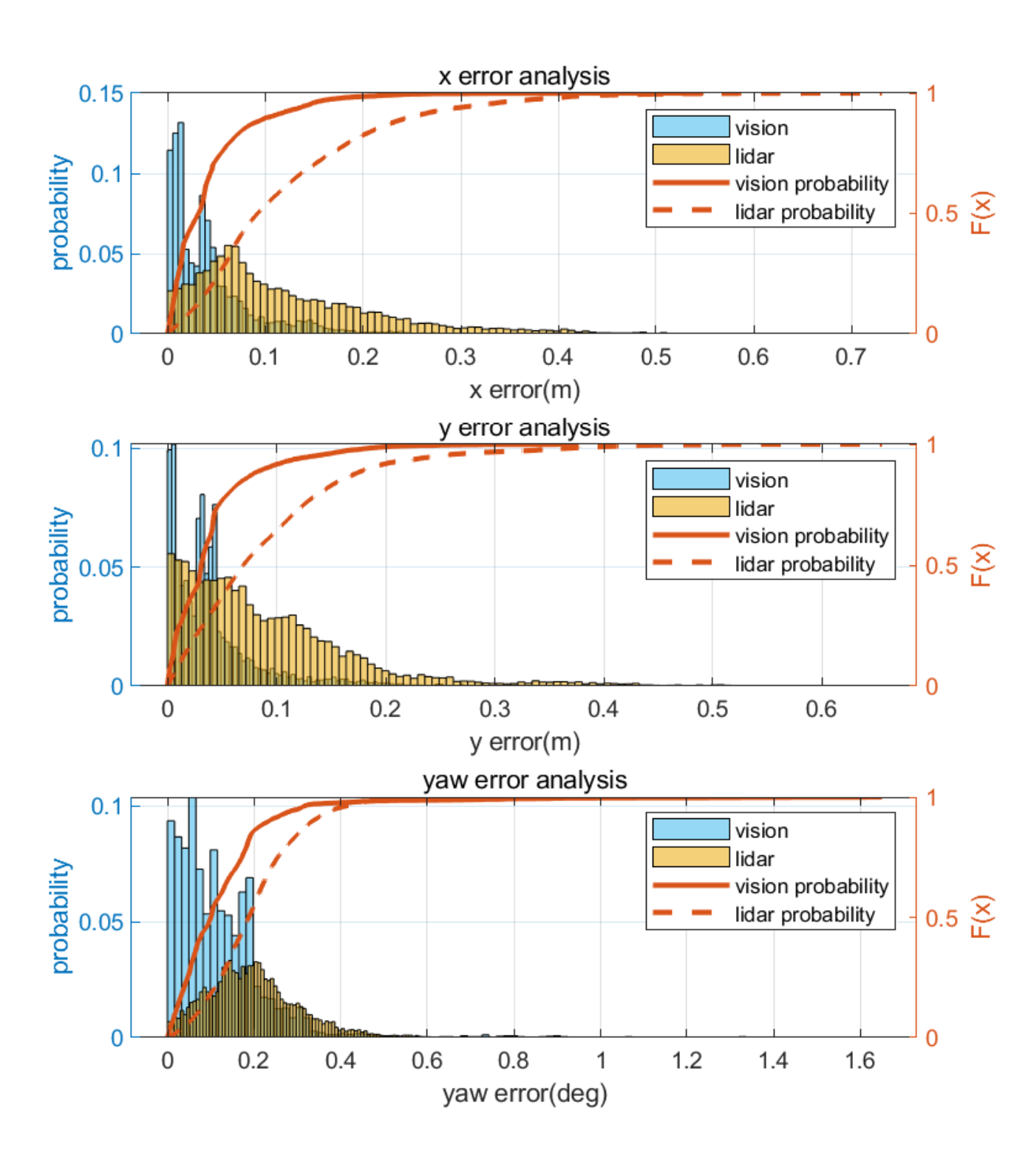}
	\caption{The probability distribution plot of localization error in x, y, and yaw (heading) directions.}
	\label{fig:8km}
\end{figure}

In this part, we evaluated the metric localization accuracy compared with the Lidar-based method.
The vehicle was equipped with a camera, Lidar, and RTK GPS.
RTK GPS was treated as the ground truth.
The semantic map produced in the first experiment was used. 
The vehicle run in this urban area.
For autonomous driving tasks, we focus on the localization accuracy on x, y directions, and the yaw (heading) angle.
Detailed results compared with Lidar were shown in Fig.\ref{fig:8km} and Table.\ref{tab:error}.
It can be seen that the proposed visual-based localization was better than the Lidar-based solution.

\begin{table}
	\centering
	\caption{{Localization Error} \label{tab:error}}
	\begin{tabular}{c|cccccc}
		\toprule
		\multirow{2}{*}{Method} & \multicolumn{2}{c}{x error {[}m{]}} & \multicolumn{2}{c}{y error {[}m{]}} & \multicolumn{2}{c}{yaw error {[}degree{]}} \\ \cline{2-7} 
		& average            & 90\%           & average            & 90\%           & average             & 90\%            \\ 
		\midrule
		vision                  & \textbf{0.043}                & \textbf{0.104}               & \textbf{0.040}                   & \textbf{0.092}               & \textbf{0.124}                    & \textbf{0.240}                
		\\
		Lidar                   & 0.121                & 0.256            & 0.091                & 0.184            & 0.197                 & 0.336             
		\\ 
		\bottomrule
	\end{tabular}
\end{table}

\section{Conclusion \& Future work}
In this paper, we proposed a novel semantic localization system, which takes full advantage of sensor-rich vehicles (e.g. Robo-taxis) to benefit low-cost production cars.
The whole framework consists of on-vehicle mapping, on-cloud updating, and user-end localization procedures.
We highlight that it is a reliable and practical localization solution for autonomous driving.

The proposed system takes advantage of markers on the road surface.
In fact, more traffic elements in 3D space can be used for localization, such as traffic light, traffic signs, and poles.
In the future, we will extend more 3D semantic features into the map.

\newpage

\balance

\bibliography{paper.bib}

\end{document}